\documentclass[11pt,a4paper]{article}
\usepackage[hyperref]{acl2019}
\usepackage{times}
\usepackage{latexsym}
\usepackage{times}
\usepackage{epsfig}
\usepackage{graphicx}
\usepackage{amsmath}
\usepackage{amssymb}
\usepackage{array}
\usepackage{multirow}
\usepackage{algpseudocode}
\usepackage{algorithm}
\usepackage{comment}
\usepackage{kotex}
\usepackage{subfig}
\usepackage{xcolor}

\usepackage{url}

\DeclareMathOperator*{\argmax}{argmax}

\aclfinalcopy 
\def\aclpaperid{460} 

\newcommand{\ds}[1]{\textcolor{black}{#1}}
\newcommand{\nj}[1]{\textcolor{black}{#1}}
\newcommand{\sh}[1]{\textcolor{black}{#1}}

\title{Textbook Question Answering with Multi-modal Context Graph Understanding and Self-supervised Open-set Comprehension}

\author{Daesik Kim$^{1,2,*}$ \quad Seonhoon Kim$^{1,3,*}$ \quad Nojun Kwak$^{1, \dagger}$ \\
$^{1}$Seoul National University \quad $^{2}$V.DO Inc. \quad $^{3}$Search\&Clova, Naver Corp.\\
\tt\small{\{daesik.kim|nojunk\}@snu.ac.kr \quad seonhoon.kim@navercorp.com}
}

\date{}

\begin{document}
\maketitle
\vspace{-3mm}
\begin{abstract}
In this work, we introduce a novel algorithm for solving the textbook question answering (TQA) task which describes more realistic QA problems compared to other recent tasks. 
We mainly focus on two related issues with analysis of the TQA dataset. 
First, \nj{solving the TQA problems} requires to comprehend multi-modal contexts in complicated input data. 
To tackle this issue of extracting knowledge features from long text lessons and merging them with visual features, we establish a context graph from texts and images, and propose a new module f-GCN based on \textit{graph convolutional networks} (GCN).
Second, scientific terms are not spread over the chapters and subjects are split in the TQA dataset. 
To overcome this so called `out-of-domain' issue, \nj{before learning QA problems,} we \nj{introduce} a novel self-supervised open-set learning process without any annotations. 
The experimental results show that our model significantly outperforms prior state-of-the-art methods. Moreover, ablation studies validate that both methods of incorporating f-GCN for extracting knowledge from multi-modal contexts and our newly proposed self-supervised learning process are \nj{effective for TQA problems}.
\end{abstract}

\section{Introduction}

{\let\thefootnote\relax\footnotetext{{
* Equal contribution. 
$\dagger$ Corresponding author.\\
This work was supported by Next-Generation Information Computing Development Program through the National Research Foundation of Korea (NRF-2017M3C4A7078547).}}}


In a decade, question answering (QA) has been one of the most promising achievements in the field of natural language processing (NLP). 
Furthermore, it has shown great potential to be applied to real-world problems.
In order to solve more realistic QA problems, input types in datasets have evolved into various combinations.
Recently, Visual Question Answering (VQA) has drawn huge attractions as it is in the intersection of vision and language.
However, the Textbook Question Answering (TQA) is a more complex \nj{and more realistic} problem as shown in Table \ref{data_type}.
Compared to context QA and VQA, the TQA uses both text and image inputs in both the context and the question.

\begin{figure}
  \includegraphics[width=\linewidth]{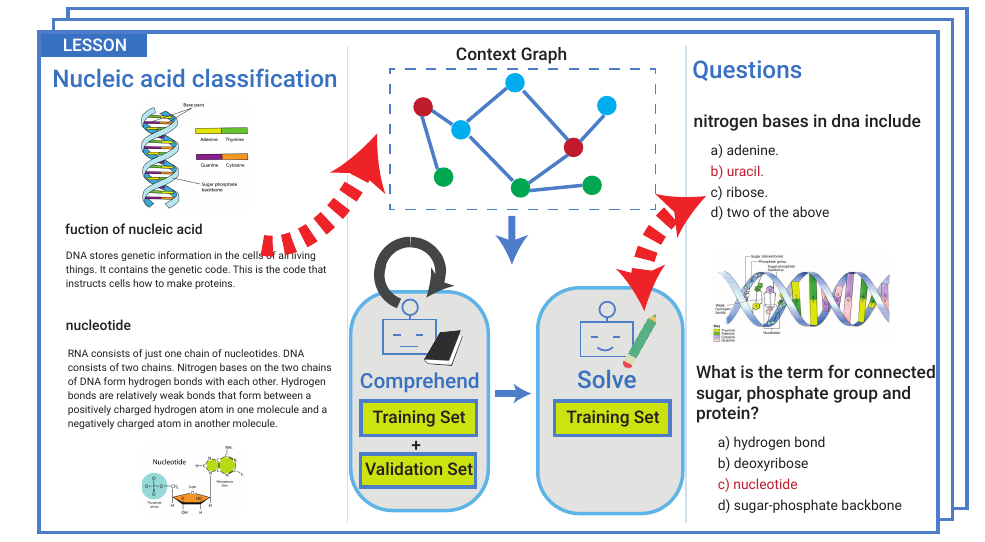}
  \caption{Examples of the \textit{textbook question answering} task and a brief concept of our work. 
  In this figure, we can see lessons which contain long essays and diagrams in the TQA. 
  Related questions are also illustrated. 
  With a self-supervised method, our model can comprehend contexts converted into context graphs in training and validation sets. Then it learns to solve questions only in the training set in a supervised manner. 
  }
  \label{overview}
\end{figure}

\begin{table}[t]
\centering
\vspace{-2mm}
\resizebox{0.9\linewidth}{!}{
\begin{tabular}{|l|c|c|c|c|c|}
\hline
\multicolumn{2}{|c|}{Input Type}  & \begin{tabular}{@{}c@{}}Context \\ QA \end{tabular}  & \begin{tabular}{@{}c@{}}Visual \\ QA \end{tabular} & \begin{tabular}{@{}c@{}}Textbook \\ QA \end{tabular}   \\  \hline
\multirow{2}{*}{Context Part} & \multicolumn{1}{c|}{Text} & \multicolumn{1}{c|}{$\circ$}& \multicolumn{1}{c|}{-}& \multicolumn{1}{c|}{$\circ$} \\\cline{2-5}
                                 & \multicolumn{1}{c|}{Image} & \multicolumn{1}{c|}{-}& \multicolumn{1}{c|}{$\circ$}& \multicolumn{1}{c|}{$\circ$} \\\hline 
\multirow{2}{*}{Question Part} & \multicolumn{1}{c|}{Text} & \multicolumn{1}{c|}{$\circ$}& \multicolumn{1}{c|}{$\circ$}& \multicolumn{1}{c|}{$\circ$} \\\cline{2-5}
                                 & \multicolumn{1}{c|}{Image} & \multicolumn{1}{c|}{-}& \multicolumn{1}{c|}{-}& \multicolumn{1}{c|}{$\circ$} \\\hline 
\end{tabular}
}
\caption{ Comparison of data types in context and question parts for context QA, VQA and TQA. It shows that the data format of the TQA task is the most complicated on both of context and question parts.}
\label{data_type}
\vspace{-3mm}
\end{table}

The TQA task can describe the real-life process of a student who learns new knowledge from books and practices to solve related problems (Figure \ref{overview}). 
It also has several novel characteristics as a realistic dataset. 
Since the TQA contains visual contents as well as textual contents, it requires to solve multi-modal QA. Moreover, formats of questions are various which include both text-related questions and diagram-related questions. 
In this paper, we focus on the following two major characteristics of the TQA dataset \cite{kembhavi2017you}.

First, compared to other QA datasets, the context part of TQA has more complexity in the aspect of data format and length.
Multi-modality of context exists even in non-diagram questions and it requires to comprehend long lessons to obtain knowledge.
Therefore, it is important to extract exact knowledge from long texts and arbitrary images.
We establish a multi-modal context graph and propose a novel module based on \textit{graph convolution networks} (GCN) \cite{kipf2016semi} to extract proper knowledge for solving questions.

Next, various topics and subjects in the textbooks are spread over chapters and lessons, and most of the knowledge and terminology do not overlap between chapters and subjects are split. 
Therefore, it is very difficult to solve problems on subjects that have not been studied before. 
To resolve this problem, we encourage our model to learn novel concepts and terms in a self-supervised manner before learning to solve specific questions. 

\nj{Our main contributions can be summarized as follows:}
\begin{itemize}
\item We propose a novel architecture which can solve TQA problems \nj{that have}  
the highest level of multi-modality.
\item We suggest a fusion GCN (f-GCN) to extract knowledge feature from the multi-modal context graph of long lessons and images in the textbook.
\item We \nj{introduce} a novel self-supervised learning process into TQA training to comprehend open-set dataset to \nj{tackle} the out-of-domain issues.
\end{itemize}
With the proposed model, we could obtain \nj{the} state-of-the-art performance on TQA dataset, which shows a large margin compared with the current state-of-the-art methods.

\begin{figure}
  \centering
  \includegraphics[width=0.9\linewidth]{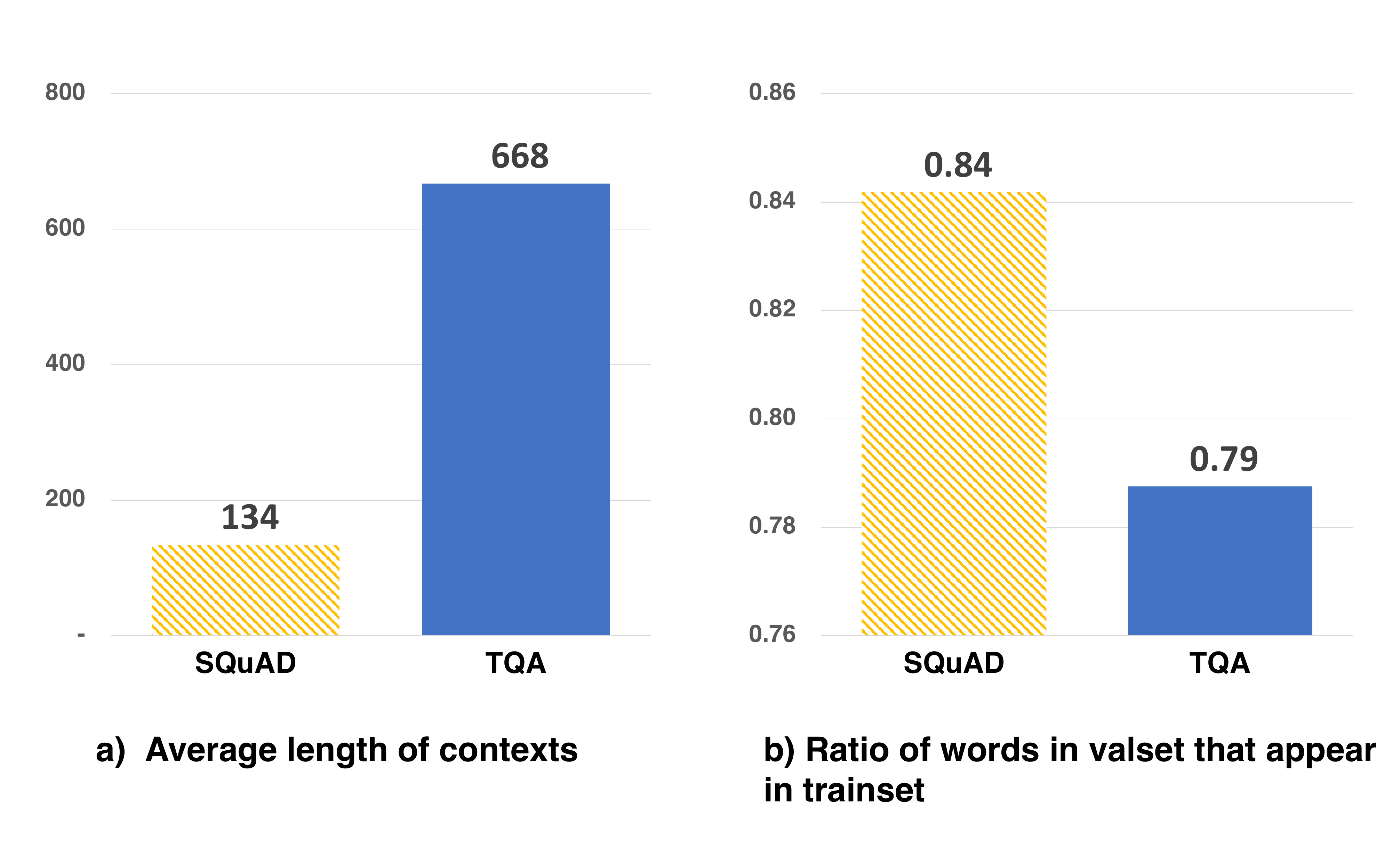}
  \vspace{-3mm}
  \caption{Analysis of contexts in TQA and SQuAD datasets. 
  }
  \label{fig:anal_context}
  \vspace{-5mm}
\end{figure}


\begin{figure*}[t]
\begin{center}
\includegraphics[width=\linewidth]{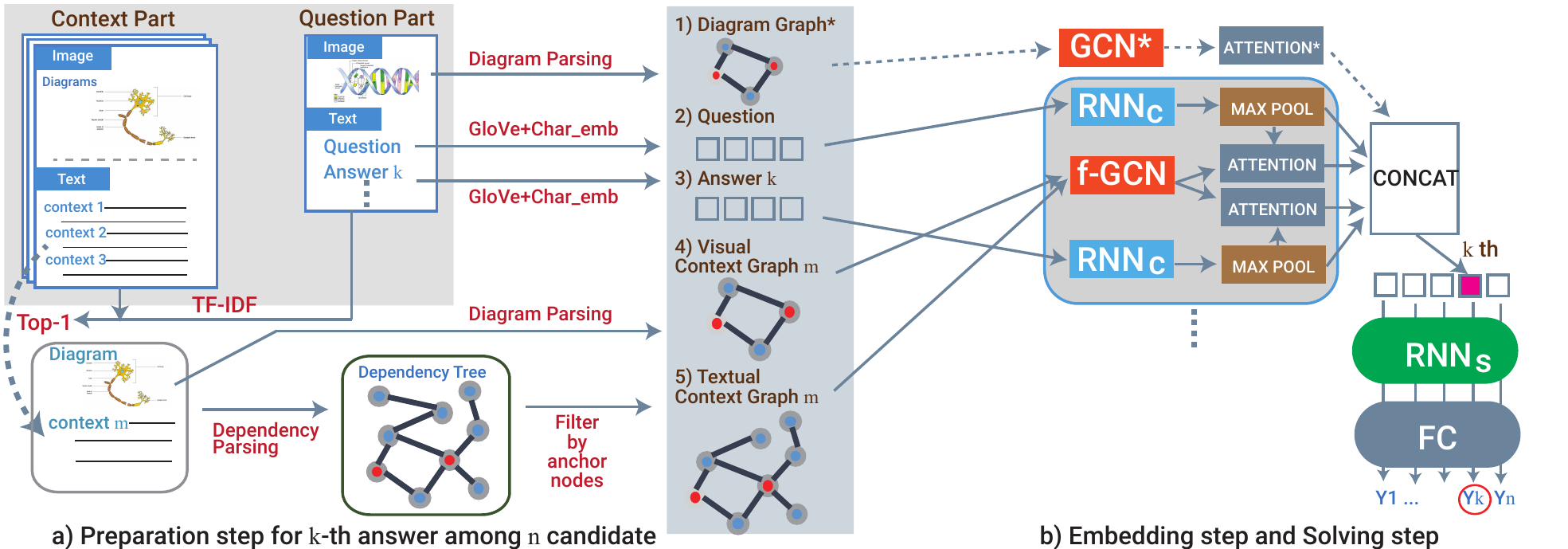}
\end{center}
\caption{\textbf{Overall framework of our model:} 
\textbf{(a) The preparation step} for the $k$-th answer among $n$ candidates. The context $m$ is determined by TF-IDF score with the question and the $k$-th answer. Then, the context $m$ is converted to a context graph $m$. The question and the $k$-th answer are also embedded by GloVe and character embedding. This step is repeated for $n$ candidates. \textbf{(b) The embedding step} uses $RNN_C$ as a sequence embedding module and f-GCN as a graph embedding module. With attention methods, we can obtain combined features. After concatenation, $RNN_S$ and the fully connected module predict final distribution in the solving step.
}
\label{fig:overall}
\vspace{-3mm}
\end{figure*}

\section{Related Work}
\label{sec:rel}
\subsection{Context question answering}
Context question answering, also known as machine reading comprehension, is a challenging task which requires a machine not only to comprehend natural language but also to reason how to answer the asked question correctly. Large amount of datasets such as MCTest \cite{richardson2013mctest}, SQuAD \cite{rajpurkar2016squad} or MS Marco \cite{nguyen2016ms} have contributed significantly to the textual reasoning via deep learning approaches. These datasets, however, are restricted to a small set of contents and contain only uni-modal problems requiring only textual information. In addition, these sets require relatively less complex parsing and reasoning compared to TQA dataset \cite{kembhavi2017you}. In this study, we tackle TQA, the practical middle school science problems across multiple modalities, by transforming long essays into customized graphs for solving the questions on a textbook.

\subsection{Visual question answering}
As the intersection of computer vision, NLP and reasoning, visual question answering has drawn attention in the last few years. Most of pioneering works in this area \cite{xu2016ask,yang2016stacked,lu2016hierarchical} are to learn a joint image-question embedding to \nj{identify} correct answers where the context is proposed by images alone. Then, various attention algorithms have been mainly  developed in this field and methods of fusing textual and visual information such as bilinear pooling \cite{fukui2016multimodal,yu2017multi} have also been widely studied. Thereafter, datasets focusing on slightly different purposes have been proposed. For instance, CLEVR \cite{johnson2017clevr} encouraged to solve the visual grounding problem and AI2D \cite{kembhavi2016diagram} suggested a new type of data for knowledge extraction from diagrams. In this paper, we incorporate UDPnet \cite{Kim_2018_CVPR} to extract knowledge from diagram parsing graph in the textbook. \ds{Recent researches \cite{teney2017graph,norcliffe2018learning} also have dealt with graph structure to solve VQA problems.}


\section{Problem}
Formally, our problem can be defined as follows: 
\begin{equation} 
\hat{a} = \argmax_{a\in \Omega_a} p(a|C,q;\theta) 
\label{eq:problem}
\end{equation}
where $C$ is given contexts which consist of textual and visual contents and $q$ is a given question which can contain question diagrams for diagram problems. $\theta$ denotes the trainable parameters. With given $C$ and $q$, we are to predict the best answer $\hat{a}$ among a set of possible answers $\Omega_a$.

The TQA contexts contain almost all items in textbooks: topic essay, diagrams and images, lesson summaries, vocabularies, and instructional videos. Among them, we mainly use topic essay as textual contexts and diagrams as visual contexts. 

Among various issues, the first problem we tackle is the complexity of contexts and variety in data formats as shown in Table \ref{data_type}. 
Especially, 
analysis of textual context in Figure \ref{fig:anal_context}(a) shows that the average length of contexts in the TQA is 668 words which is almost 5 times larger than that of the SQuAD which has 134 words on average.
Also, in \cite{kembhavi2017you}, analysis of information scope in TQA dataset provides two important clues that about 80\% of text questions only need 1 paragraph and about 80\% of diagram questions only need 1 context image and 1 paragraph.
Due to those evidences, we need to add an information retrieval step such as TF-IDF (term frequency--inverse document frequency) to narrow down scope of contexts from a lesson to a paragraph, \nj{which significantly reduces the} complexity of a problem.
Moreover, a graph structure can be suitable to represent logical relations between scientific terms and to merge them with visual contexts from diagrams.
As a result, we decide to build a multi-modal context graph and obtain knowledge features from it.

In Figure \ref{fig:anal_context}(b), we obtain the percentage of how much the terms in the validation set are appearing in the training set. 
Obviously, the ratio of the TQA \nj{(79\%)} is lower than that of the SQuAD \nj{(84\%)} which can induce out-of-vocabulary and domain problems more seriously in the TQA task. To avoid aforementioned issues, we apply a novel self-supervised learning process before learning to solve questions.

\section{Proposed Method}
Figure \ref{fig:overall} illustrates our overall framework which consists of three steps. 
In a preparation step, we use TF-IDF to select the paragraph most relevant to the given question or candidate answers. 
Then, we convert it into two types of context graphs for text and image, respectively.
In the embedding step, we exploit an RNN (denoted as RNN$_C$ in the figure) to embed textual inputs, a question and an answer candidate. 
Then, we incorporate f-GCN to extract graph features from both the visual and the textual context graphs.
After repeating previous steps for each answer candidate, we can stack each of concatenated features from the embedding step. 
\ds{We exploit another RNN (RNN$_S$) to \nj{cope with the} variable \nj{number of answer candidates} which \nj{varies} from 2 to 7 \nj{that}} \sh{can} \ds{have sequential relations such as ``none of the above"} 
\sh{or}
\ds{``all of the above" in \nj{the} last choice. }
Final fully connected layers decide probabilities of answer candidates. 
\ds{Note that notation policies are included in \nj{the} supplementary.}


\subsection{Multi-modal Context Graph Understanding}


\subsubsection{Visual and Textual Context graphs}

For the visual contexts and the question diagrams, we build a visual context graph using UDPnet \cite{Kim_2018_CVPR}. 
We obtain names\sh{, counts, and relations of entities in diagrams.}
Then we can establish edges between related entities. 
Only for question diagrams, we use counts of entities transformed in the form of a sentence such as ``There are 5 objects" or ``There are 6 stages".

We build the textual context graphs using some parts of the lesson where the questions can focus on solving problems as follows. 
Each lesson can be divided into multiple paragraphs and we extract one paragraph which has the highest TF-IDF score using a concatenation of the question and one of the candidate answers (leftmost of Figure \ref{fig:overall}(a)). 

Then, we build the dependency trees of the extracted paragraph utilizing the Stanford dependency parser \cite{manning2014stanford}, and designate the words which exist in the question and the candidate answer as anchor nodes. 
The nodes which have more than two levels of depth difference with anchor nodes are removed and we build the textual context graphs using the remaining nodes and edges (Process 1 in the supplementary).



\begin{figure}[t]
\begin{center}
\includegraphics[width=\linewidth]{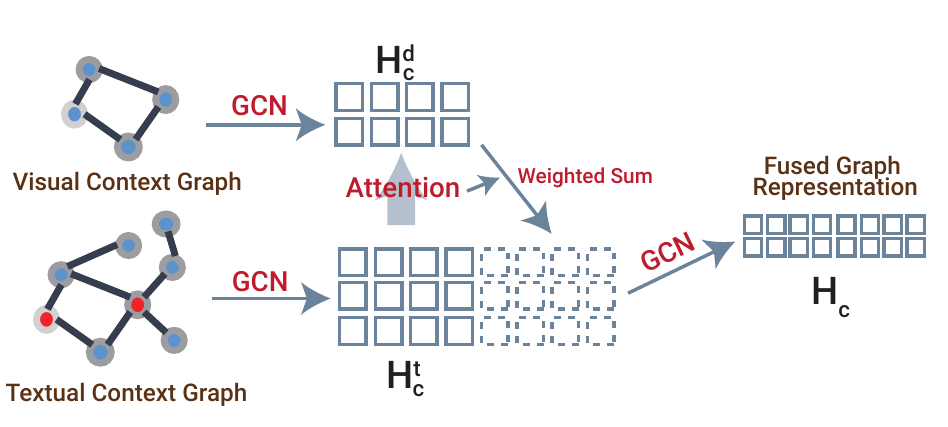}
\end{center}
\vspace{-3mm}
\caption{Illustration of f-GCN. Both of textual and visual contexts are converted into $H_c^d$ and $H_c^t$.
With attention methods, we obtain combined features of $H_c^t$ and $H_c^d$ (f-GCN1).
Finally, we use GCN again to propagate over entire features of context graphs (f-GCN2).
}
\label{fig:fgcn}
\vspace{-3mm}
\end{figure}

\subsubsection{Graph Understanding using f-GCN}

Next, 
we propose f-GCN to extract combined graph features for visual and textual context graphs as shown in Figure \ref{fig:fgcn}.
Each of context graphs has its own graph matrix $C$ containing node features and a normalized adjacency matrix \sh{which} are used as inputs of a GCN to comprehend the contexts. Here, the graph matrix $C$ is composed of the word embeddings and the character representation.
First, we extract propagated graph features from both of context graphs based on one-layer GCN as
\begin{equation} 
\begin{split}
H_c^t=&f(C^t,\mathcal{A}^t)= \sigma(\mathcal{A}^tC^tW^t) \\
H_c^d=&f(C^d,\mathcal{A}^d)= \sigma(\mathcal{A}^dC^dW^d),
\end{split}
\label{eq:H}
\end{equation}
where 
$\mathcal{A}^t$ and $\mathcal{A}^d$ are the adjacency matrices for the text and visual contexts,
$W^t$ and $W^d$ are learning parameters of linear layer for the text and visual contexts, and the element-wise operation $\sigma$ is the $\tanh$ 
activation function. 


After that, we 
\sh{use dot product function to}
get attention matrix $Z$ of visual context $H_c^d$ against textual context $H_c^t$ which contains main knowledge. Then we concatenate features of textual context $H_c^t$ and weighted sum $Z^TH_c^d$ to get entire context features, 
\begin{equation} 
H_c^1=[H_c^t;Z^TH_c^d],\
\label{eq:H1}
\end{equation}
where  $[\cdot \ ; \ \cdot]$ is the concatenation operator.
Compared to the textual-context-only case, we can obtain double-sized features which can be more informative. 
Finally, we use a GCN again to propagate over entire features of context graphs:
\begin{equation} 
\begin{split}
H_c^2=&f(H_c^1,\mathcal{A}^t)= \sigma(\mathcal{A}^tH_c^1 W^c) .
\end{split}
\label{eq:H2}
\end{equation}
We denote this module except the last GCN as f-GCN1 \nj{(eq. (\ref{eq:H1}))} and the whole module including the last GCN as f-GCN2 \nj{(eq. (\ref{eq:H2}))}.

\subsection{Multi-modal Problem Solving}
The f-GCN and RNNs are used to embed the contexts and answer the questions as shown in Figure \ref{fig:overall}(b). Two different RNNs are used in our architecture. One is the \textit{comprehending} RNN (RNN$_{C}$) which can understand questions and candidate answers and the other is the \textit{solving} RNN (RNN$_{S}$) which can answer the questions. 

The input of the RNN$_{C}$ is comprised of the word embedding, character representation and the occurrence flag for both questions and candidate answers. In word embedding, each word can be represented as $e_{q_i}$/$e_{a_i}$ by using a pre-trained word embedding method such as GloVe \cite{pennington2014glove}. The character representation $c_{q_i}$/$c_{a_i}$ is calculated by feeding randomly initialized character embeddings into a CNN with the max-pooling operation. The occurrence flag $f_{q_i}$/$f_{a_i}$ indicates whether the word occurs in the contexts or not. Our final input representation $q_i^w$ for the question word $q_i$ in RNN$_{C}$ is composed of three components as follows:
\begin{equation}
\label{eq:input_rnn_c}
\begin{split}
e_{q_i}=& Emb(q_i), \quad
c_{q_i} = \textit{Char-CNN}(q_i) \\
&q_i^w = [e_{q_i}; c_{q_i}; f_{q_i}].
\end{split}
\end{equation}
The input representation for the candidate answers is also obtained in the same way as the one for the question.
Here, $Emb$ is the trainable word embeddings and  \textit{Char-CNN} is the character-level convolutional network. To extract proper representations for the questions and candidate answers, we apply the step-wise max-pooling operation over the RNN$_C$ hidden features.

Given each of the question and the candidate answer representations, we use an attention mechanism to focus on the relevant parts of the contexts for solving the problem correctly. The attentive information $Att_q$ of the question representation $h_q$ against the context features $H_c$ as in (\ref{eq:H1}) or (\ref{eq:H2}) is calculated as follows:
\begin{equation}
\begin{split}
Att_q = \sum_{k=1}^{K} \alpha_k H_{c_k}, & \quad
\alpha_k = \frac{exp(g_k)}{\sum_{i=1}^{K}exp(g_i)}, \\
g_k =  & \;  h_q^T\mathbf{M}H_{c_k}.
\end{split}
\label{eq:attention}
\end{equation}
Here, $K$ is the number of words in the context $C$ which equals the dimension of the square adjacency matrix $\mathcal{A}$.
$\mathbf{M}$ is the attention matrix that converts the question into the context space. The attentive information of the candidate answers $Att_a$ is calculated similar to $Att_q$.

RNN$_{S}$ can solve the problems and its input consists of the representations of the question and the candidate answer with their attentive information on the contexts as:
\begin{equation}
\begin{split}
I_{RNN_S}^t &= [h_q; h_a; Att_q^c; Att_a^c], \\
I_{RNN_S}^d &= [h_q; h_a; Att_q^c; Att_a^c; Att_q^{qd}; Att_a^{qd}]
\end{split}
\label{eq:s_rnn_input2}
\end{equation}
where 
$I_{RNN_S}^t$ is for the text questions and 
$I_{RNN_S}^d$ is for the diagram questions.
Finally, based on the outputs of RNN$_{S}$, we use one fully-connected layer followed by a softmax function to obtain a probability distribution of each candidate answer and optimize those with cross-entropy loss.

\begin{figure}[t]
\begin{center}
\includegraphics[width=\linewidth]{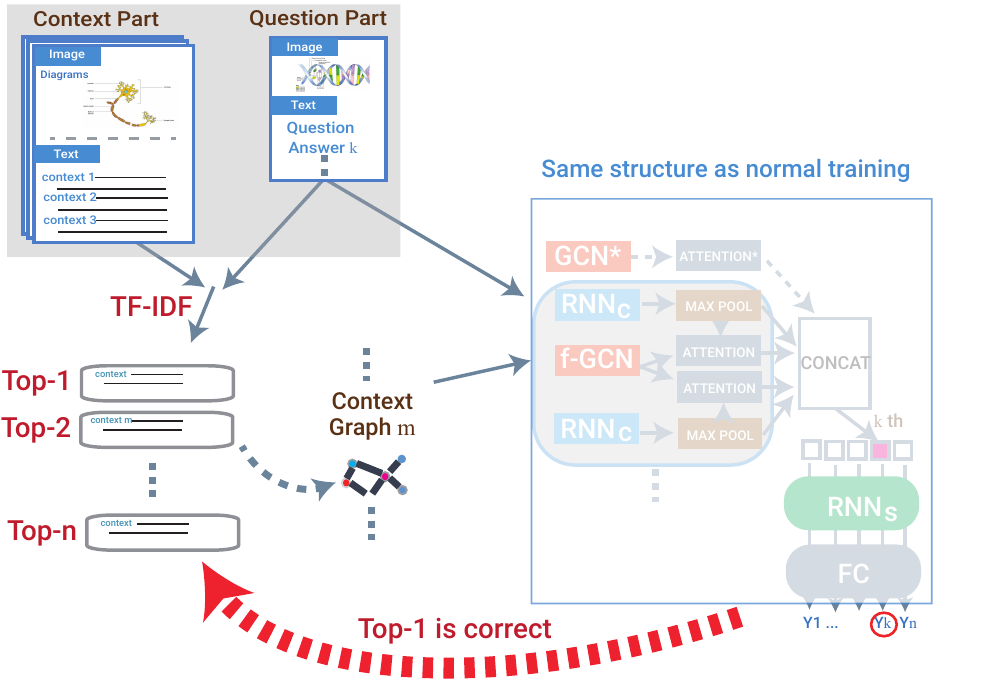}
\vspace{-3mm}
\end{center}
\vspace{-3mm}
\caption{Self-supervised open-set comprehension step in our model. 
We set contexts as candidates we should predict for the question and the $k$-th answer. For each answer, we obtain $n$ context candidates from TF-IDF methods and set the top-1 candidate as the correct context. While we use the same structure as in Figure \ref{fig:overall}, we can predict final distribution after all the steps.  }
\label{fig:long}
\vspace{-3mm}
\end{figure}

\subsection{Self-supervised open-set comprehension}

To comprehend out-of-domain contexts, we propose a self-supervised prior learning method as shown in Figure \ref{fig:long}. 
While we exploit the same architecture described in the previous section, 
we have reversed the role of the candidate answer and the contexts in (\ref{eq:problem}) as a self-supervised one. In other words, we set the problem as inferring the Top-1 context for the chosen answer candidate. We assume TF-IDF to be quite reliable in measuring closeness between texts.

The newly defined self-supervised problem can be formalized as follows: 
\begin{equation} 
\hat{c} = \argmax_{c\in \Omega_c} p(c|A_k,q;\theta) 
\end{equation}
where $A_k$ is given $k$-th answer candidate among $n$ candidates and $q$ is the given question. 
Then we infer the most related context $\hat{c}$ among a set of contexts $\Omega_c$ in a lesson.

\begin{table*}[t]
\centering
\resizebox{0.7\textwidth}{!}{
\begin{tabular}{l c c c c c}
\hline\hline
Model  & Text T/F  & Text MC  & Text All  & Diagram & All   \\  \hline
Random & 50.10 & 22.88 & 33.62 & 24.96 & 29.08 \\ 
MemN+VQA \cite{kembhavi2017you} & 50.50 & 31.05 & 38.73 & 31.82 & 35.11 \\
MemN+DPG \cite{kembhavi2017you} & 50.50 & 30.98 & 38.69 & 32.83 & 35.62 \\ 
BiDAF+DPG \cite{kembhavi2017you} & 50.40 & 30.46 & 38.33 & 32.72 & 35.39 \\ 
Challenge & - & - & 45.57 & 35.85 & 40.48   \\
IGMN \cite{li2018textbook} & 57.41 & 40.00 & 46.88 & 36.35 & 41.36 \\  \hline
Our full model  w/o visual context				& 62.32	&49.15	&54.35	&36.61	&45.06  \\ 
Our full model  w/ f-GCN2				& 62.22	&48.76	&54.11	&\textbf{37.72}	&45.52  \\ 
Our full model   				& \textbf{62.73}	&\textbf{49.54}	&\textbf{54.75}	&37.61	&\textbf{45.77}  \\ 
\hspace{3mm} w/o SSOC(VAL)   	& 62.22	&48.82	&54.11	&37.47	&45.39  \\ 
\hspace{3mm} w/o SSOC(TR+VAL)  	& 60.02	&46.86	&52.06	&36.61 &43.97  \\ 
\hspace{3mm} w/o f-GCN $\And$ SSOC(TR+VAL)  & 58.72 & 45.16 & 50.51 & 35.67 & 42.74   \\   


\hline \hline
\end{tabular}
}
\caption{ Comparison of performance with previous methods (Top) and results of ablation studies (Bottom). We demonstrate the accuracies of each type of questions, Text T/F (true-false in text only), Text MC (multiple-choices in text only), Text all (all in text only), Diagram and All. 
Note that previous methods only used textual context.
}
\label{exp1}
\vspace{-3mm}
\end{table*}

\ds{
For each candidate answer $A_k (k=1,..,n)$, we get the set of paragraphs $\Omega_c$ of size $j$ from the corresponding context. 
Here, $\Omega_c$ is obtained by calculating TF-IDF between $[q;A_k]$ and each paragraph $\omega$, i.e.,  $T_{\omega} =\textit{tf-idf}([q;A_k],\omega)$, and  
selecting the top-$j$ paragraphs.
Among the $j$ paragraphs $\omega_i (i=1, \cdots, j)$ in $\Omega_c$, the one with the highest TF-IDF score is set as the ground truth:
\vspace{-2mm}
\begin{equation}
\label{eq:uoc}
\begin{split}
y_i = \begin{cases}
  1, & \text{if } \omega_i = \argmax_{\omega\in \Omega_c}T_{\omega}, \\
  0, & \text{otherwise}.
\end{cases}
\end{split}
\end{equation}
With $A_k$, $q$ and $\omega_i \in \Omega_c$, we conduct the same process in eq. (2-7) to obtain the $i$-th input of the $RNN_S$, ${I_{RNN_S}^i}$.
After repeating it $j$ times, we put all $I_{RNN_S}^i, (i=1\cdots, j)$ into $RNN_S$ sequentially and optimize this step with the cross-entropy loss.
We repeatedly choose all answer candidates $A_k$, and conduct the same process in this step.  
}

With this pre-training stage which shares parameters with the supervised stage, we expect that our model can deal with almost all contexts in a lesson. 
Moreover, it becomes possible to learn contexts in the validation set or the test set with a self-supervised manner. 
This step is analogous to a student who reads and understands a textbook and problems in advance.

\section{Experiments}
\subsection{Dataset}

We perform experiments on the TQA dataset, which consists of 1,076 lessons from Life Science, Earth Science and Physical Science textbooks. 
While the dataset contains 78,338 sentences and 3,455 images including diagrams, it also has 26,260 questions with 12,567 of them having an accompanying diagram, split
into training, validation and test at a lesson level.
The training set consists of 666 lessons and 15,154 questions, the validation set consists of 200 lessons and 5,309 questions and the test set consists of 210 lessons and 5,797 questions. Since evaluation for test is hidden, we only use the validation set to evaluate our methods.

\subsection{Baselines}
We compare our method with several recent methods
as followings:

\noindent $\bullet$ \textbf{MemN+VQA, MemN+DPG}
Both exploits Memory networks to embed texts in lessons and questions. First method uses VQA 
\sh{approaches} for diagram questions, and 
the second one exploits Diagram Parse Graph (DPG) as context graph on diagrams built by DsDP-net \cite{kembhavi2016diagram}.


\noindent $\bullet$  \textbf{BiDAF+DPG} 
It incorporates BiDAF (Bi-directional Attention Flow Network) \cite{seo2016bidirectional}, a recent machine comprehension model which exploits a bidirectional attention mechanism to capture dependencies between question and corresponding context paragraph. 

For above 3 models, we use experimental results newly reported in \cite{li2018textbook}.

\noindent $\bullet$  \textbf{Challenge} This is the one that obtained the top results in TQA competition \cite{kembhavi2017you}. 
The results in the table are mixed with each of top score in \sh{the} text-question track and the diagram-question track.

\noindent $\bullet$  \textbf{IGMN} It uses the Instructor Guidance with Memory Net
s (IGMN) based on Contradiction Entity-Relationship Graph (CERG). For diagram questions, it only recognizes texts in diagrams.

\noindent $\bullet$ \textbf{Our full model w/o visual context} This method excludes visual context to compare with previous methods on the same condition. It uses only one-layer GCN for textual context and self-supervised open-set comprehension (SSOC).

\noindent $\bullet$ \textbf{Our full model w/ f-GCN2} From now, all methods include visual context. This method uses f-GCN2 and SSOC. 

Following methods are for our ablation study:

\noindent $\bullet$  \textbf{Our full model} This method uses both of our methods, f-GCN1 and SSOC on the training and the validation sets.

\noindent $\bullet$  \textbf{Our model w/o SSOC (VAL)} This method only uses training set to pretrain parameters in SSOC.

\noindent $\bullet$  \textbf{Our model w/o SSOC (TR+VAL)} This method eliminates whole SSOC pre-training process. It only uses f-GCN as Graph extractor and was trained only in a normal supervised learning manner.

\noindent $\bullet$  \textbf{Our model w/o f-GCN$\And$SSOC (TR+VAL)} This method ablates both f-GCN module and SSOC process. It replaces f-GCN as vanilla RNN, other conditions are the same.

\subsection{Quantitative Results}
\subsubsection{Comparison of Results}
Overall results on TQA dataset are shown in Table \ref{exp1}. 
The results show that all variants of our model outperform other recent models in all type of question. Our best model shows about 4\% higher than state-of-the-art model in overall accuracy. Especially, an accuracy in text question significantly outperforms other results with about 8\% margin. 
A result on diagram questions also shows more than 1\% increase over the previous best model. 
We believe that our two novel proposals, context graph understanding and self-supervised open-set comprehension work well on this problem since our models achieve significant margins compared to recent researches. 


Even though our model w/o visual context only uses one-layer GCN for textual context, it shows better result compared to MemN+VQA and MemN+DPG with a large margin and IGMN with about 3\% margin. 
IGMN also exploits a graph module of contraction, but ours outperforms especially in both text problems, T/F and MC with over 5\% margin. 
We believe that the graph in our method can directly represents the feature of context and the GCN also plays an important role in extracting the features of our graph. 

Our models with multi-modal contexts show significantly better results on both text and diagram questions. Especially, results of diagram question outperform over 1\% rather than our model w/o visual context. 
Those results indicate that f-GCN sufficiently exploits visual contexts to solve diagram questions.

\subsubsection{Ablation Study}
We perform ablation experiments in Table \ref{exp1}.
Our full model w/ f-GCN2 can achieve best score on diagram questions but slightly lower scores on text questions. Since the overall result of our full model records the best, we conduct ablation study of each module of it.

\begin{table}[t]
\vspace{-2mm}
\centering
\resizebox{0.8\linewidth}{!}{
\begin{tabular}{l c c c c c}
\hline\hline
Model    & Text  & Diagram & All   \\  \hline

Our model w/o SSOC  	& \textbf{52.06} & \textbf{36.61} & \textbf{43.97}  \\ 
\hspace{4mm} w/o q-flag    	& 49.29 & 35.78 & 42.21  \\ 
\hspace{4mm} w/o a-flag   	& 43.24 & 31.50 & 37.09  \\ 
\hspace{4mm} w/o q$\And$a-flag  & 42.64 & 31.72 & 36.92   \\   
\hline \hline
\end{tabular}
}
\caption{Results of ablation study about the occurrence flags. We demonstrate the accuracies of Text only, 
Diagram, 
and total questions without SSOC method.}
\label{exp2}
\vspace{-3mm}
\end{table}

\begin{figure*}[t]
\begin{center}
  \includegraphics[width=\linewidth]{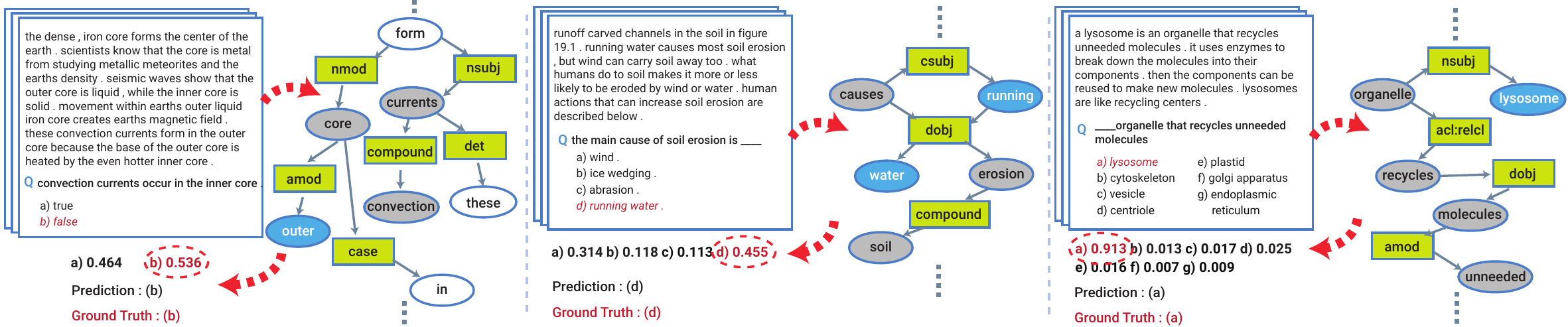}
  \end{center}
  \caption{Qualitative results of text-type questions without visual context. 
  Each example shows all items for a question in the textbook and a textual context subgraph to solve a question. And our predicted distribution for answers and ground truths are also displayed. In the subgraph, gray circles represent words in questions and blue circles represent words related to answers. Green rectangles represent relation types of the dependency graph.}
  \label{fig:qual}
  \vspace{-3mm}
\end{figure*}

\begin{figure}[t]
\begin{center}
\includegraphics[width=0.96\linewidth]{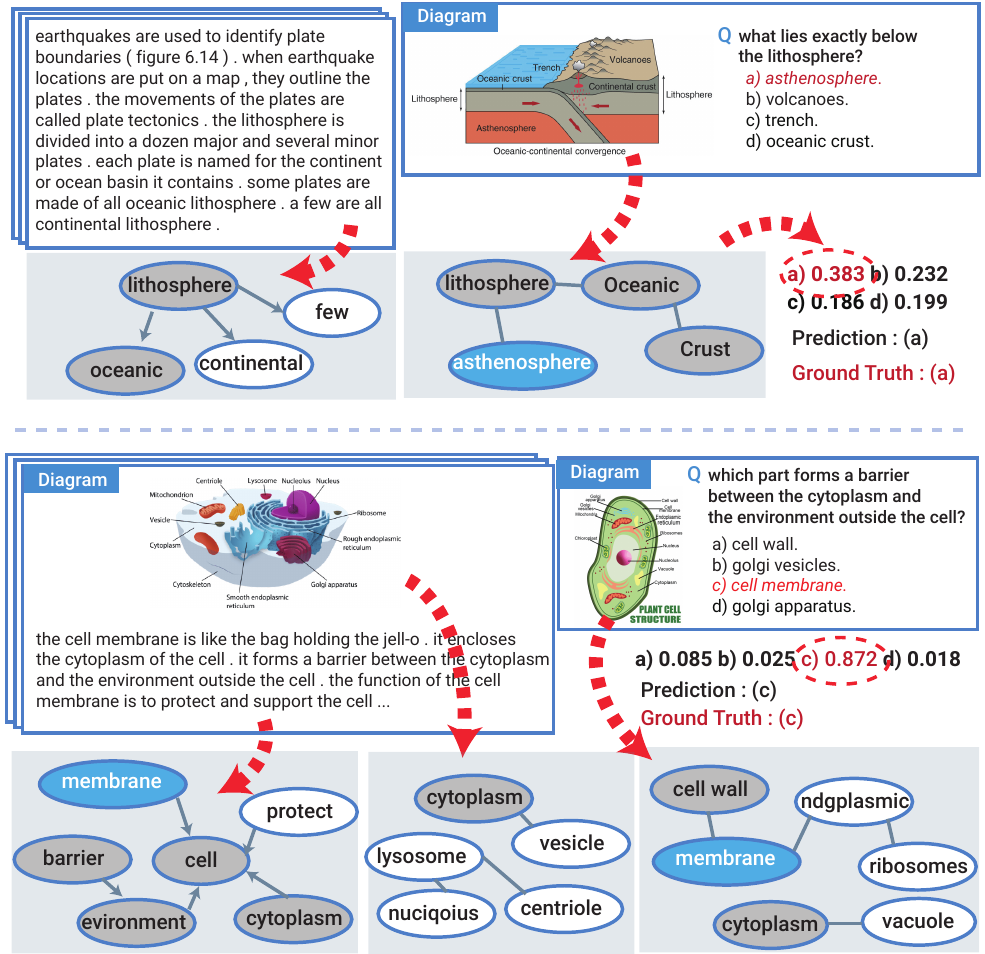}
\end{center}
\vspace{-3mm}
\caption{Qualitative results of diagram-type questions. We illustrate intermediate subgraphs, and predicted distribution for answers and ground truths.}
\vspace{-3mm}
\label{fig:qual2}
\vspace{-3mm}
\end{figure}

First, we observe an apparent decrease in our model when any part of modules is eliminated. 
It is surprising that self-supervised open-set comprehension method provides an improvement on our model. 
Our full model shows about 2\% higher performance than the model without SSOC(TR+VAL). 
It is also interesting to compare our full model with our model without SSOC(VAL). 
The results show that using the additional validation set on SSOC can improve overall accuracy compared to using only training set. 
It seems to have more advantage for learning unknown dataset in advance.  

Our model without f-GCN$\And$SSOC eliminates our two novel modules and replace GCN with vanilla RNN. 
That model shows 1\% of performance degradation compared with the model without SSOC(TR+VAL) which means that it might not sufficient to deal with knowledge features with only RNN and attention module. 
Thus, context graph we create for each lesson could give proper representations with f-GCN module.

Table \ref{exp2} shows the results of ablation study about occurrence flag. All models do not use SSOC method. 
In (\ref{eq:input_rnn_c}), we concatenate three components including the occurrence flag to create question or answer representation. 
We found that the occurrence flag which explicitly indicates the existence of a corresponding word in the contexts has a meaningful effect. 
Results of all types degrade significantly as ablating occurrence flags. Especially, eliminating a-flag drops accuracy about 7\% which is almost 4 times higher than the decrease due to eliminating f-flag. 
We believe that disentangled features of answer candidates can mainly determine the results while a question feature equally affects all features of candidates. Our model without both flags shows the lowest results due to the loss of representational power.

\subsection{Qualitative Results}

Figure \ref{fig:qual} shows three qualitative results of text-type questions without visual context.
We illustrate textual contexts, questions, answer candidates and related subgraphs of context graphs. 

The first example describes a pipeline on a T/F question. Three words, ``currents", ``core" and ``convection" are set as anchor nodes as shown in the left of Figure \ref{fig:qual}. Within two levels of depth, we can find ``outer" node which is the opposite to ``inner" in the question sentence. As a result, our model predicts the true and false probabilities of this question as 0.464 and 0.536, respectively, and correctly solves this problem as a false statement.
Next example is a multiple choice problem which is more complicated than T/F problem. With anchor nodes which consist of each answer candidate and a question such as ``causes", ``erosion" and ``soil", the context graph can be established including nodes in two depth of graph from anchor nodes. Among the 4 candidates, choice (d) contains the same words, ``running" and ``water", as our model predicts. Therefore, our model can estimate (d) as the correct answer with the highest probability of 0.455. 
The last example shows a more complicated multiple choice problem. In the context graph, we set ``organelle", ``recycles", ``molecules" and ``unneeded" as anchor nodes with each word in answer candidates. Then we can easily find an important term, ``lysosome" in choice (a). Therfore, choice (a) has a probability close to one among 7 candidates. 

Figure \ref{fig:qual2} demonstrates qualitative results of diagram questions. 
We exclude relation type nodes in subgraphs of the dependency tree for simplicity and also illustrate diagram parsing graphs of visual contexts and question diagram.
The example in the top shows intermediate results of subgraphs on a diagram question without visual context.
Even though chosen paragraph in textual context do not include ``asthenosphere", graph of a question diagram contain relation between ``asthenosphere" and ``lithosphere". Then our model can predict (a) as the correct answer with probability of 0.383.
The bottom illustration describes the most complex case which has diagrams in both of context and question parts. 
We illustrate all subgraphs of text and diagrams. 
While our model can collect sufficient knowledge about cell structure on broad information scope, ``cell membrane" can be chosen as correct answer with the highest probability.

These examples demonstrate abstraction ability and relationship expressiveness which can be huge advantages of graphs. 
Moreover, those results could support that our model can explicitly interpret the process of solving multi-modal QA.

\section{Conclusion}
In this paper, we proposed two novel methods to solve a realistic task, TQA dataset.
We extract knowledge features with \sh{the} proposed f-GCN and conduct self-supervised learning to overcome the out-of-domain issue.
Our method also demonstrates state-of-the-art results.
We believe that our work can be a meaningful step in realistic multi-modal QA and solving the out-of-domain issue.

{\small
\bibliographystyle{acl_natbib}
\bibliography{acl2019}

\begin{thebibliography}{19}
\expandafter\ifx\csname natexlab\endcsname\relax\def\natexlab#1{#1}\fi

\bibitem[{Fukui et~al.(2016)Fukui, Park, Yang, Rohrbach, Darrell, and
  Rohrbach}]{fukui2016multimodal}
Akira Fukui, Dong~Huk Park, Daylen Yang, Anna Rohrbach, Trevor Darrell, and
  Marcus Rohrbach. 2016.
\newblock Multimodal compact bilinear pooling for visual question answering and
  visual grounding.
\newblock \emph{arXiv preprint arXiv:1606.01847}.

\bibitem[{Johnson et~al.(2017)Johnson, Hariharan, van~der Maaten, Fei-Fei,
  Zitnick, and Girshick}]{johnson2017clevr}
Justin Johnson, Bharath Hariharan, Laurens van~der Maaten, Li~Fei-Fei,
  C~Lawrence Zitnick, and Ross Girshick. 2017.
\newblock Clevr: A diagnostic dataset for compositional language and elementary
  visual reasoning.
\newblock In \emph{Computer Vision and Pattern Recognition (CVPR), 2017 IEEE
  Conference on}, pages 1988--1997. IEEE.

\bibitem[{Kembhavi et~al.(2016)Kembhavi, Salvato, Kolve, Seo, Hajishirzi, and
  Farhadi}]{kembhavi2016diagram}
Aniruddha Kembhavi, Mike Salvato, Eric Kolve, Minjoon Seo, Hannaneh Hajishirzi,
  and Ali Farhadi. 2016.
\newblock A diagram is worth a dozen images.
\newblock In \emph{European Conference on Computer Vision}, pages 235--251.
  Springer.

\bibitem[{Kembhavi et~al.(2017)Kembhavi, Seo, Schwenk, Choi, Farhadi, and
  Hajishirzi}]{kembhavi2017you}
Aniruddha Kembhavi, Minjoon Seo, Dustin Schwenk, Jonghyun Choi, Ali Farhadi,
  and Hannaneh Hajishirzi. 2017.
\newblock Are you smarter than a sixth grader? textbook question answering for
  multimodal machine comprehension.
\newblock In \emph{2017 IEEE Conference on Computer Vision and Pattern
  Recognition (CVPR)}, pages 5376--5384. IEEE.

\bibitem[{Kim et~al.(2018)Kim, Yoo, Kim, Lee, and Kwak}]{Kim_2018_CVPR}
Daesik Kim, YoungJoon Yoo, Jee-Soo Kim, SangKuk Lee, and Nojun Kwak. 2018.
\newblock Dynamic graph generation network: Generating relational knowledge
  from diagrams.
\newblock In \emph{The IEEE Conference on Computer Vision and Pattern
  Recognition (CVPR)}.

\bibitem[{Kipf and Welling(2016)}]{kipf2016semi}
Thomas~N Kipf and Max Welling. 2016.
\newblock Semi-supervised classification with graph convolutional networks.
\newblock \emph{arXiv preprint arXiv:1609.02907}.

\bibitem[{Li et~al.(2018)Li, Su, Zhu, Wang, and Zhang}]{li2018textbook}
Juzheng Li, Hang Su, Jun Zhu, Siyu Wang, and Bo~Zhang. 2018.
\newblock Textbook question answering under instructor guidance with memory
  networks.
\newblock In \emph{Proceedings of the IEEE Conference on Computer Vision and
  Pattern Recognition}, pages 3655--3663.

\bibitem[{Lu et~al.(2016)Lu, Yang, Batra, and Parikh}]{lu2016hierarchical}
Jiasen Lu, Jianwei Yang, Dhruv Batra, and Devi Parikh. 2016.
\newblock Hierarchical question-image co-attention for visual question
  answering.
\newblock In \emph{Advances In Neural Information Processing Systems}, pages
  289--297.

\bibitem[{Manning et~al.(2014)Manning, Surdeanu, Bauer, Finkel, Bethard, and
  McClosky}]{manning2014stanford}
Christopher Manning, Mihai Surdeanu, John Bauer, Jenny Finkel, Steven Bethard,
  and David McClosky. 2014.
\newblock The stanford corenlp natural language processing toolkit.
\newblock In \emph{Proceedings of 52nd annual meeting of the association for
  computational linguistics: system demonstrations}, pages 55--60.

\bibitem[{Nguyen et~al.(2016)Nguyen, Rosenberg, Song, Gao, Tiwary, Majumder,
  and Deng}]{nguyen2016ms}
Tri Nguyen, Mir Rosenberg, Xia Song, Jianfeng Gao, Saurabh Tiwary, Rangan
  Majumder, and Li~Deng. 2016.
\newblock Ms marco: A human generated machine reading comprehension dataset.
\newblock \emph{arXiv preprint arXiv:1611.09268}.

\bibitem[{Norcliffe-Brown et~al.(2018)Norcliffe-Brown, Vafeias, and
  Parisot}]{norcliffe2018learning}
Will Norcliffe-Brown, Stathis Vafeias, and Sarah Parisot. 2018.
\newblock Learning conditioned graph structures for interpretable visual
  question answering.
\newblock In \emph{Advances in Neural Information Processing Systems}, pages
  8344--8353.

\bibitem[{Pennington et~al.(2014)Pennington, Socher, and
  Manning}]{pennington2014glove}
Jeffrey Pennington, Richard Socher, and Christopher Manning. 2014.
\newblock Glove: Global vectors for word representation.
\newblock In \emph{Proceedings of the 2014 conference on empirical methods in
  natural language processing (EMNLP)}, pages 1532--1543.

\bibitem[{Rajpurkar et~al.(2016)Rajpurkar, Zhang, Lopyrev, and
  Liang}]{rajpurkar2016squad}
Pranav Rajpurkar, Jian Zhang, Konstantin Lopyrev, and Percy Liang. 2016.
\newblock Squad: 100,000+ questions for machine comprehension of text.
\newblock \emph{arXiv preprint arXiv:1606.05250}.

\bibitem[{Richardson et~al.(2013)Richardson, Burges, and
  Renshaw}]{richardson2013mctest}
Matthew Richardson, Christopher~JC Burges, and Erin Renshaw. 2013.
\newblock Mctest: A challenge dataset for the open-domain machine comprehension
  of text.
\newblock In \emph{Proceedings of the 2013 Conference on Empirical Methods in
  Natural Language Processing}, pages 193--203.

\bibitem[{Seo et~al.(2016)Seo, Kembhavi, Farhadi, and
  Hajishirzi}]{seo2016bidirectional}
Minjoon Seo, Aniruddha Kembhavi, Ali Farhadi, and Hannaneh Hajishirzi. 2016.
\newblock Bidirectional attention flow for machine comprehension.
\newblock \emph{arXiv preprint arXiv:1611.01603}.

\bibitem[{Teney et~al.(2017)Teney, Liu, and van~den Hengel}]{teney2017graph}
Damien Teney, Lingqiao Liu, and Anton van~den Hengel. 2017.
\newblock Graph-structured representations for visual question answering.
\newblock In \emph{Proceedings of the IEEE Conference on Computer Vision and
  Pattern Recognition}, pages 1--9.

\bibitem[{Xu and Saenko(2016)}]{xu2016ask}
Huijuan Xu and Kate Saenko. 2016.
\newblock Ask, attend and answer: Exploring question-guided spatial attention
  for visual question answering.
\newblock In \emph{European Conference on Computer Vision}, pages 451--466.
  Springer.

\bibitem[{Yang et~al.(2016)Yang, He, Gao, Deng, and Smola}]{yang2016stacked}
Zichao Yang, Xiaodong He, Jianfeng Gao, Li~Deng, and Alex Smola. 2016.
\newblock Stacked attention networks for image question answering.
\newblock In \emph{Proceedings of the IEEE Conference on Computer Vision and
  Pattern Recognition}, pages 21--29.

\bibitem[{Yu et~al.()Yu, Yu, Fan, and Tao}]{yu2017multi}
Zhou Yu, Jun Yu, Jianping Fan, and Dacheng Tao.
\newblock Multi-modal factorized bilinear pooling with co-attention learning
  for visual question answering.

\end{thebibliography}
}

\appendix

\begin{table*}[t]
\centering
\resizebox{0.8\textwidth}{!}{
\begin{tabular}{l c c c c c}
\hline\hline
Model  & Text T/F  & Text MC  & Text All  & Diagram & All   \\  \hline
Our full model w/o visual context				& \textbf{62.32}	&\textbf{49.15}	&\textbf{54.35}	&\textbf{36.61}	&\textbf{45.06}  \\ 
\hspace{3mm} w/o UTC(VAL)   	& 60.82	&49.08	&53.72	&36.53	&44.72  \\ 
\hspace{3mm} w/o UTC(TR+VAL)  	& 60.72	&46.34	&52.02	&36.57 &43.93  \\ 
\hspace{3mm} w/o GCN $\And$ UTC(TR+VAL)  & 58.62 & 44.77 & 50.24 & 35.2 & 42.36   \\  \hline
Our full model w/ f-GCN2				& \textbf{62.22}	&\textbf{48.76}	&\textbf{54.11}	&\textbf{37.72}	&\textbf{45.52}  \\ 
\hspace{3mm} w/o UTC(VAL)   	& 62.63	&48.43	&54.03	&37.32	&45.28  \\ 
\hspace{3mm} w/o UTC(TR+VAL)  	& 61.42	&46.67	&52.49	&36.71 &44.22  \\ 
\hspace{3mm} w/o GCN $\And$ UTC(TR+VAL)  & 58.72 & 45.16 & 50.51 & 35.67 & 42.74   \\  

\hline \hline
\end{tabular}
}
\caption{ Results of additional ablation studies. We demonstrate the accuracies of each type of questions: Text T/F (true-false in text only), Text MC (multiple-choices in text only), Text all (all in text only), Diagram and All (total questions). Results of our full model \nj{without} visual context are on the top of the table and results of our full model \nj{with} f-GCN2 are in the bottom. 
}
\label{tab:ablation}
\end{table*}

\section{Notations}

We denote the question text, question diagram, candidate answer, text context and diagram context as 
$Q^t = \{q^t_{1}, q^t_{2}, \cdots, q^t_{I}\}$, 
$Q^d = \{q^d_{1}, q^d_{2}, \cdots, q^d_{J}\}$,
$A = \{a_{1}, a_{2}, \cdots, a_{K}\}$,
$C^t = \{c^t_{1}, c^t_{2}, \cdots, c^t_{L}\}$,
and $C^d = \{c^d_{1}, c^d_{2}, \cdots, c^d_{M}\}$, 
respectively where $q^t_i$/$q^d_j$/$a_k$/$c^t_l$/$c^d_m$ is the $i^{th}$/$j^{th}$/$k^{th}$/$l^{th}$/$m^{th}$ word of the question text $Q^t$ and the question diagram $Q^d$, candidate answer $A$, text context $C^t$ and diagram context $C^d$ ($C$ is unified notation for the $C^t$ and $C^d$).
The corresponding representations are denoted as $h_q^t$,$h_q^d$, $h_a$, $H_c^t$ and $H_c^d$, respectively. 
Note that we use the diagram context $C^d$ only in the diagram questions.

\section{Implementation Details}

We initialized word embedding with 300d GloVe vectors pre-trained from the 840B Common Crawl corpus, while the word embeddings for the out-of-vocabulary words were initialized randomly. We also randomly initialized character embedding with a 16d vector and extracted 32d character representation with a \sh{1D} convolutional network. \sh{And the 1D convolution kernel size is 5.} We used \sh{200} hidden units of Bi-LSTM for the RNN$_c$ whose weights are shared between the question and \nj{the} candidate answers. The maximum sequence length of them is \nj{set} to 30. Likewise, the number of hidden units of the RNN$_s$ is the same as the RNN$_c$ and the maximum sequence length is 7 which is the same as the number of the \nj{maximum} candidate answers. 
\sh{We employed 200d one layer GCN for all types of graphs, and the number of maximum nodes is 75 for the textual context graph, 35 for the diagrammatic context graph, and 25 for the diagrammatic question graph, respectively.}
We use \nj{$\tanh$} for the activation function of the GCN. The dropout was applied after all of the word embeddings with a keep rate of 0.5. The Adam optimizer with an initial learning rate of 0.001 was applied, and the learning rate was decreased by a factor of 0.9 after each epoch.

\begin{figure}[h]
\begin{center}
\includegraphics[width=\linewidth]{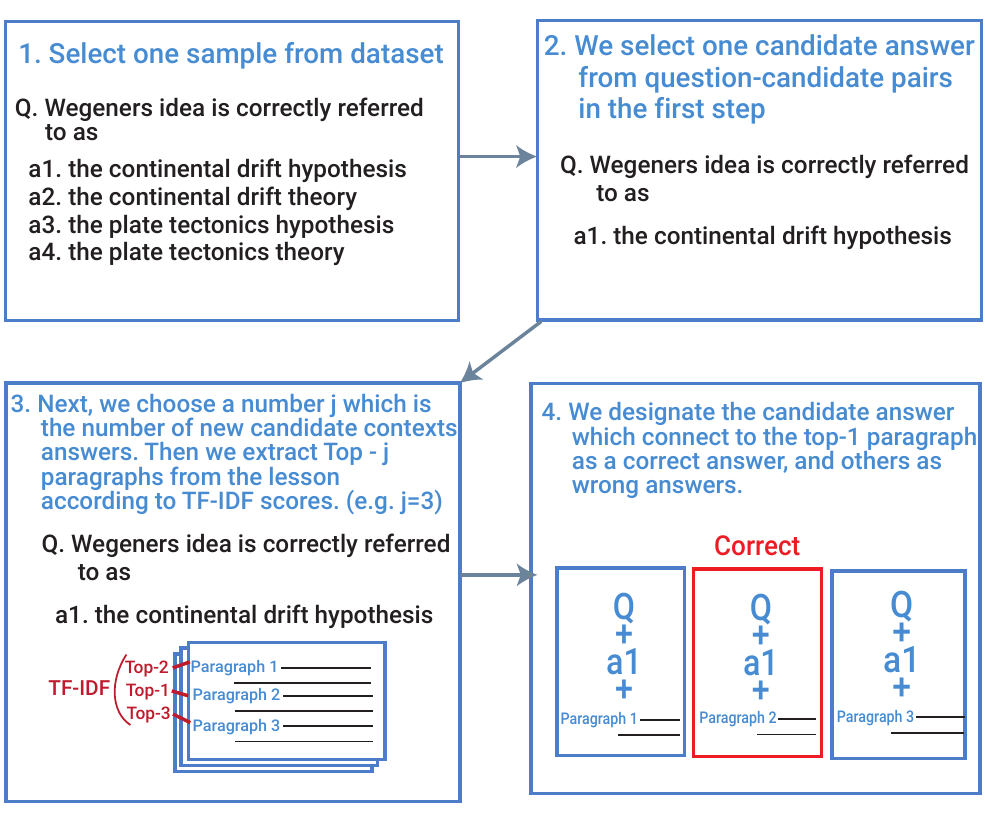}
\end{center}
   \caption{Additional examples of SSOC steps.}
\label{fig:sup0}
\end{figure}

\section{Additional explanation for SSOC}

In Figure \ref{fig:sup0}, we illustrate examples about detailed steps of SSOC.
\sh{In the first step}, we select one candidate answer from question-candidate answers pairs 
(2). Next, we choose a number $j$, the number of candidate contexts for the pair of question-candidate answer, in the range 2 to 7 like the original dataset (3). If $j$ is higher than the number of contexts in the lesson, we set $j$ to be the number of contexts. Then, we extract top $j$ paragraphs using the TF-IDF scores to set them as candidate contexts $\Omega_c$ (3). We build each context graph in the same way as the original method and get embeddings with the question-candidate answer pair we selected. Finally, we designate the final candidate which connects to the top 1 paragraph as a correct answer, and others as wrong answers (4).

\section{Results of additional ablation study}

We perform additional ablation studies for variants of our model.
For both \nj{o}ur full model \nj{without} visual context and \nj{our} full model \nj{with} f-GCN2, results of ablation studies are shown in Table \ref{tab:ablation}.
Both studies seem to demonstrate similar tendency as performances are degraded for ablating each module.
We can conclude that our two novel modules have sufficient contributions to improve \nj{the performance} our model in the TQA problem.

\section{Process of Building \sh{Textual} Context Graph}

{
\makeatletter
\renewcommand{\ALG@name}{Process}
\makeatother
\begin{algorithm}[h]
\caption{Build \ds{textual} context and adjacency matrices $C$, $\mathcal{A}$}\label{alg:euclid}
 \hspace*{\algorithmicindent} 
 	\textbf{Input:} 
    	a paragraph, 
        a set of \textit{anchor nodes} $V$
\begin{algorithmic}[1]
\State Construct a dependency tree on each sentence 
       of the given paragraph
\State Split the tree into multiple units each of which represents two nodes and one edge $u=\{v_1,v_2\}$
\State $U\gets$ a set of units
\State $E\gets$ an empty set of edges
\For{$depth\gets 1$ to $2$}
  	\For{all nodes $v \in V$}
	  	\For{all units $u \in U$}
        	\If{$v \in u$}
            	\State $E\gets E \cup \{u\}$
            \EndIf
  		\EndFor
  	\EndFor
	\State $V\gets$ a set of all nodes in $E$
\EndFor
\end{algorithmic}
 \hspace*{\algorithmicindent} \textbf{Output:} 
 	context matrix $C$ from $V$ with embedding matrices, 
    adjacency matrix $\mathcal{A}$ from $E$
\end{algorithm}
}

\sh{The procedure for converting the textual context into the graph structures is shown in Process 1. After constructing the dependency trees, we set the nodes included in the question or the candidate answer as anchor nodes and built the final context graph $C$ by removing the nodes which have more than two levels of depth difference with anchor nodes. We also constructed the adjacency matrix $\mathcal{A}$ using the remaining nodes and edges.}

\section{Additional Qualitative Results}

\begin{figure*}[t]
\begin{center}
\includegraphics[width=\textwidth]{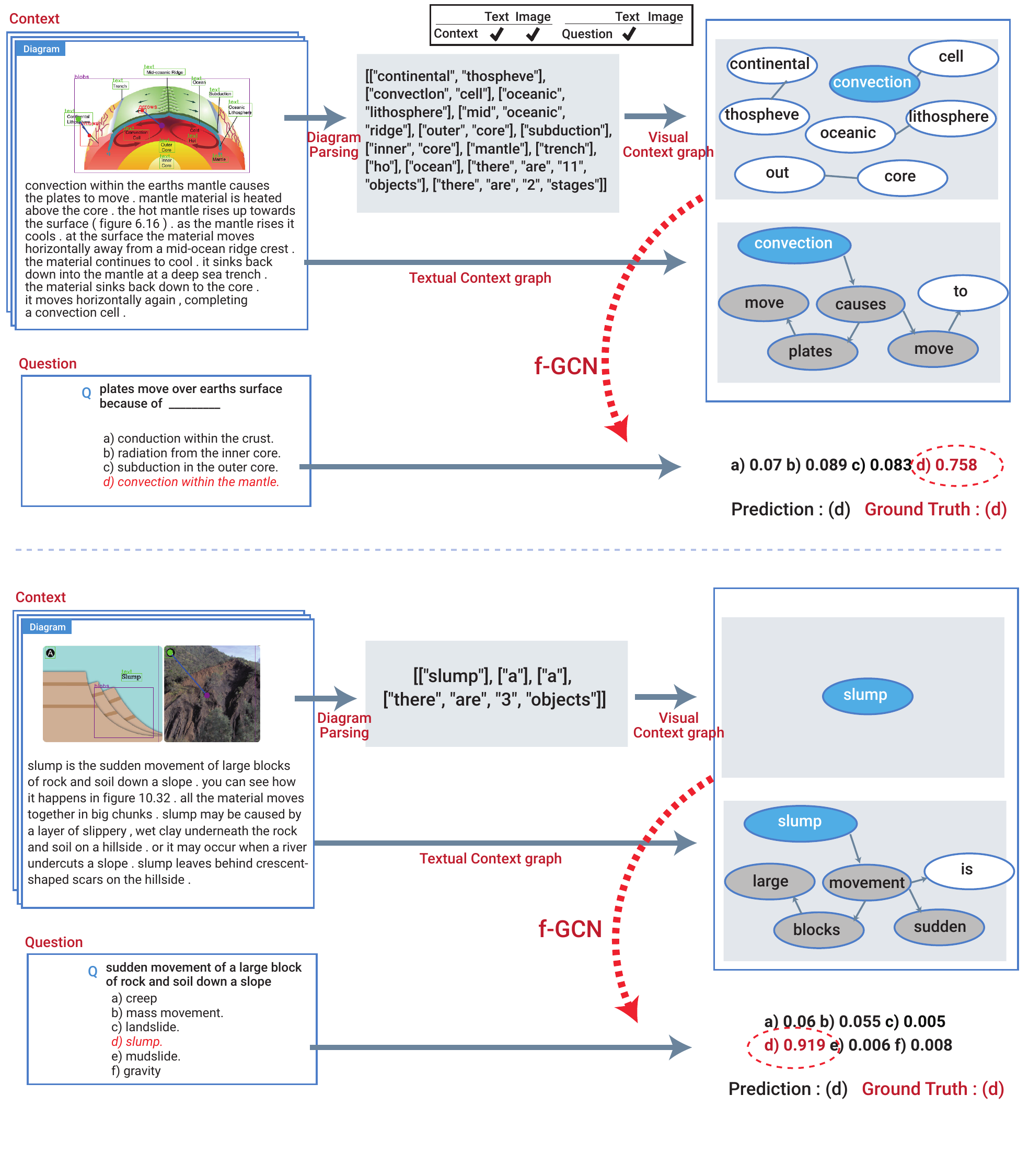}
\end{center}
   \caption{Additional qualitative results on text-type question with visual context. For both examples, a pipeline from visual context to visual context graph is shown. Gray circles represent words in questions and blue circles represent words related to answers. }
\label{fig:sup1}
\end{figure*}

\begin{figure*}[t]
\begin{center}
\includegraphics[width=\textwidth]{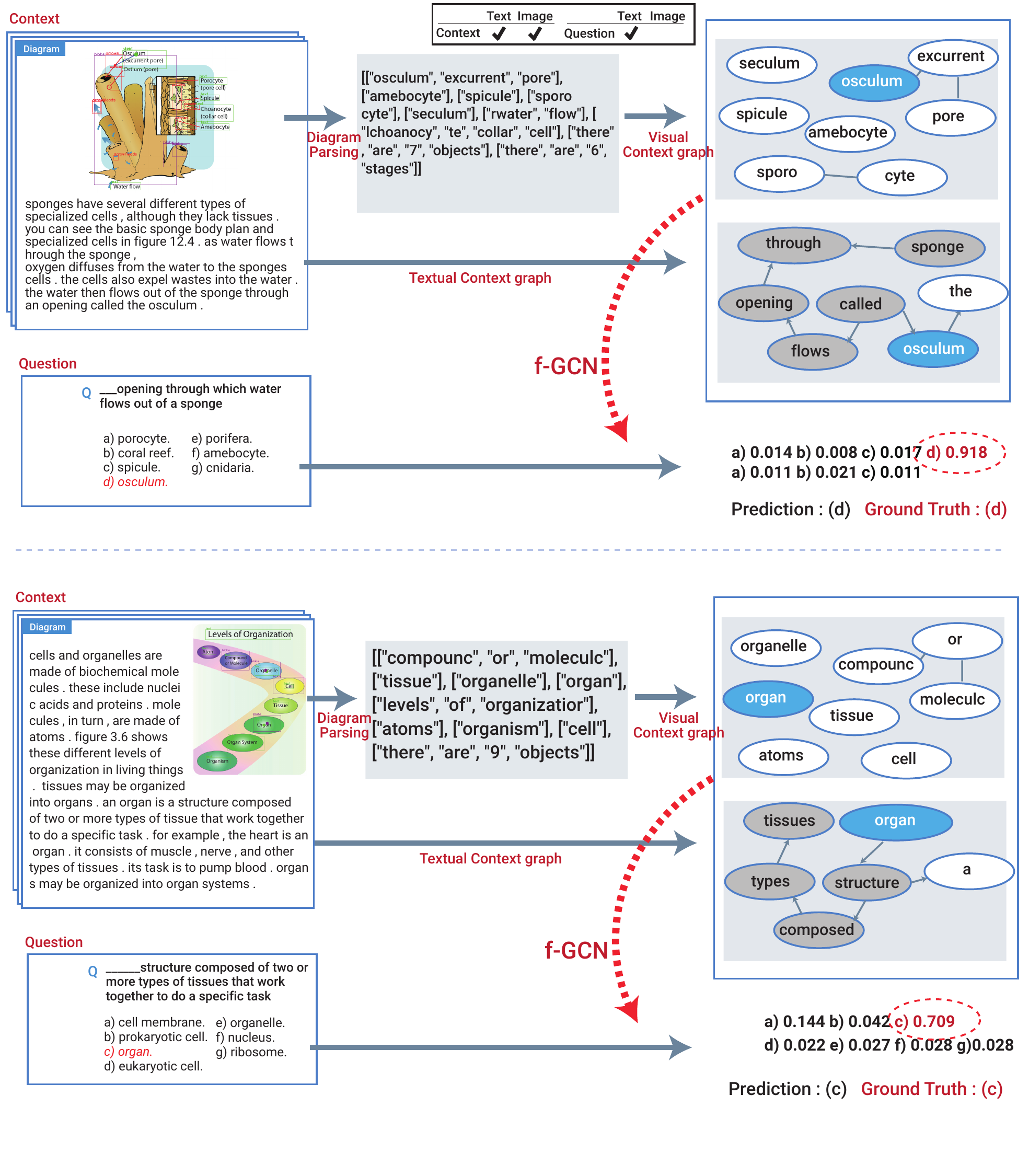}
\end{center}
   \caption{Additional qualitative results on text-type question with visual context. For both examples, a pipeline from visual context to visual context graph is shown. Gray circles represent words in questions and blue circles represent words related to answers.}
\label{fig:sup2}
\end{figure*}

\begin{figure*}[t]
\begin{center}
\includegraphics[width=\textwidth]{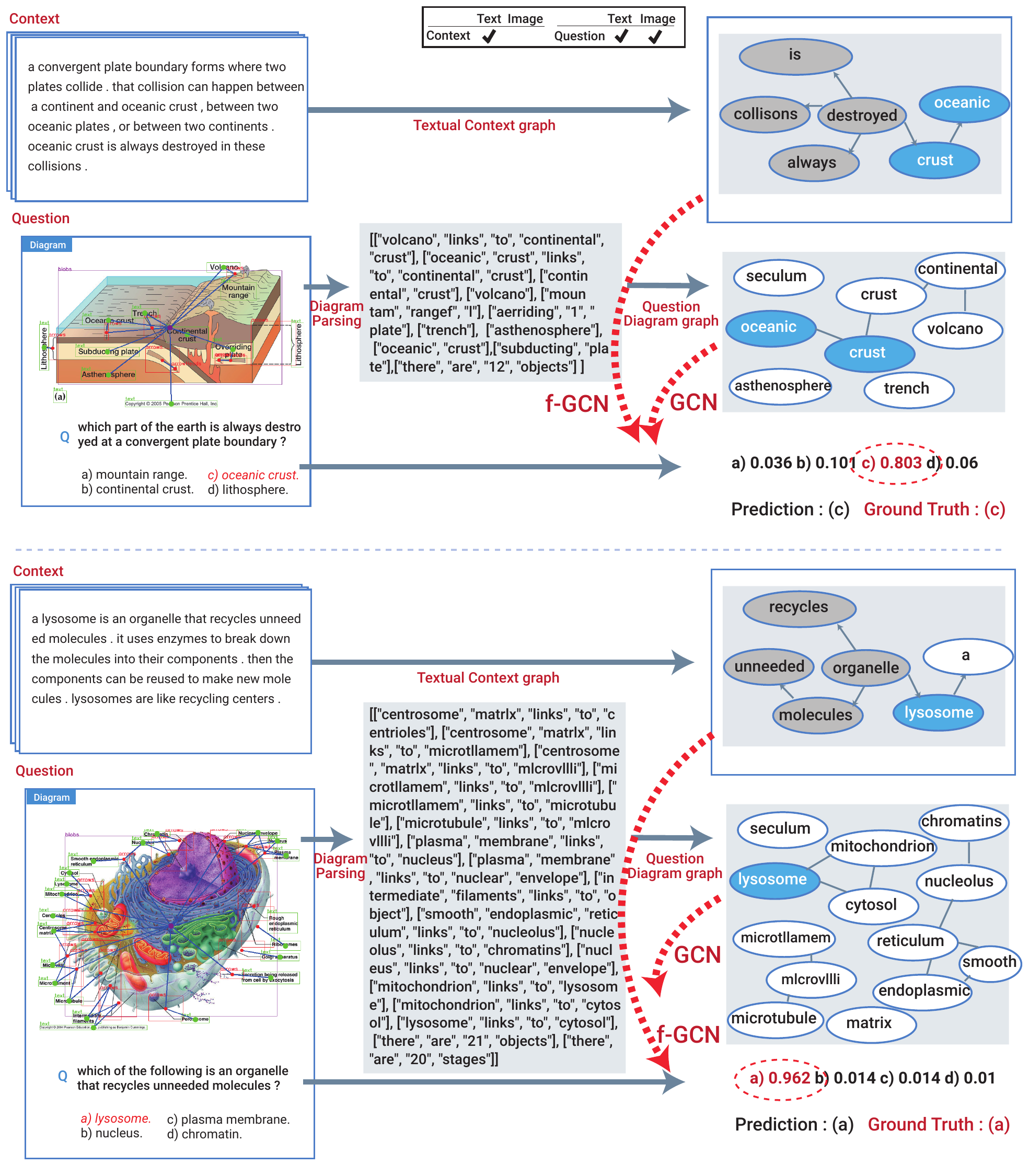}
\end{center}
   \caption{Additional qualitative results on diagram-type question without visual context. For both examples, a pipeline from question diagram to question diagram graph is shown. Gray circles represent words in questions and blue circles represent words related to answers.}
\label{fig:sup3}
\end{figure*}

\begin{figure*}[t]
\begin{center}
\includegraphics[width=\textwidth]{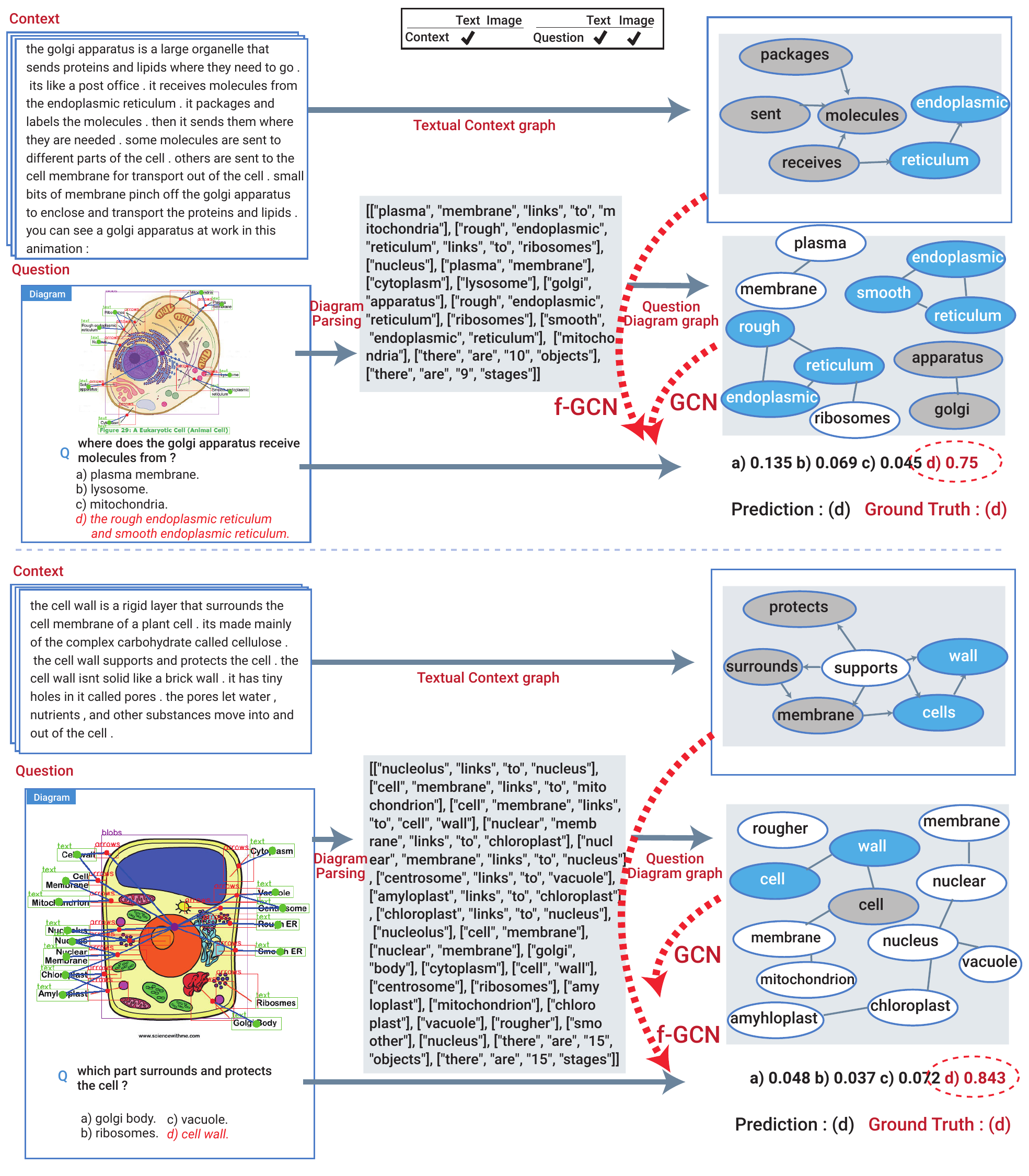}
\end{center}
   \caption{Additional qualitative results on diagram-type question without visual context. For both examples, a pipeline from question diagram to question diagram graph is shown. Gray circles represent words in questions and blue circles represent words related to answers.}
\label{fig:sup4}
\end{figure*}

\begin{figure*}[t]
\begin{center}
\includegraphics[width=0.95\textwidth]{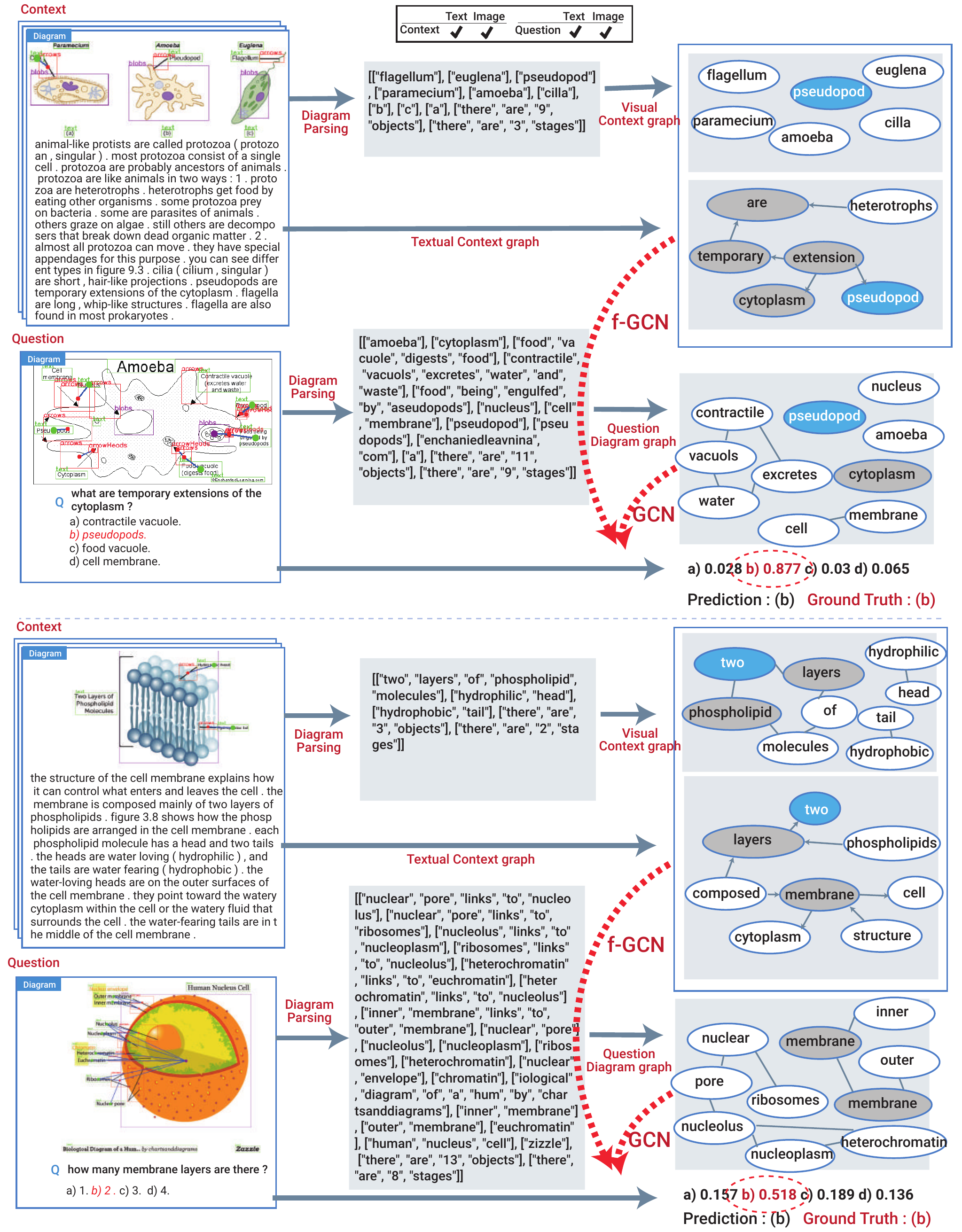}
\end{center}
   \caption{Additional qualitative results on diagram-type question with visual context. For both examples, pipelines from visual context and question diagram to visual context graph and question diagram graph are shown. Gray circles represent words in questions and blue circles represent words related to answers.}
\label{fig:sup5}
\end{figure*}

\begin{figure*}[t]
\begin{center}
\includegraphics[width=0.95\textwidth]{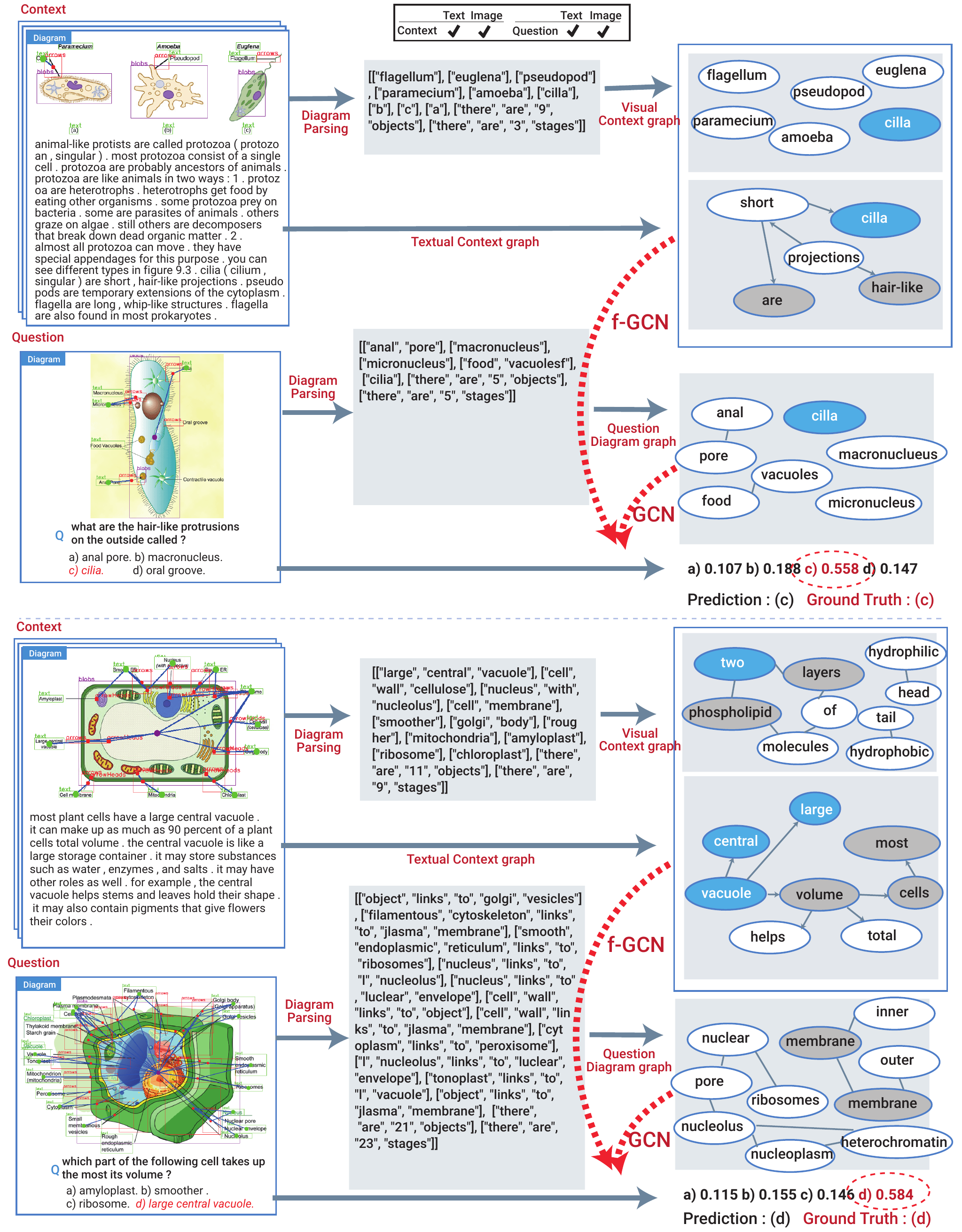}
\end{center}
   \caption{Additional qualitative results on diagram-type question with visual context. For both examples, pipelines from visual context and question diagram to visual context graph and question diagram graph are shown. Gray circles represent words in questions and blue circles represent words related to answers.}
\label{fig:sup6}
\end{figure*}

In next pages, we present additional qualitative results of questions in three types. We explicitly demonstrates all intermediate results as subgraphs of visual context and question diagram. Note that we add a legend that indicates which types of data are used in this figure to avoid confusion. In Figure \ref{fig:sup1} and Figure \ref{fig:sup2}, we illustrate intermediate and final results on text-type question with visual context. Next, we demonstrate intermediate and final results on diagram-type question without visual context in Figure \ref{fig:sup3} and Figure \ref{fig:sup4}. Finally, we present intermediate and final results of  the most complicated type, diagram-type question with visual context in Figure \ref{fig:sup5} and Figure \ref{fig:sup6}. We hope the logical connectivity for solving the problem and how our model works well on the TQA problem are sufficiently understood with those figures. 


\end{document}